\documentstyle[12pt,endnotes,lingmacros]{article}
\let\footnote=\endnote

\newcommand{\bit}{\begin{itemize}}
\newcommand{\eit}{\end{itemize}}
\newcommand{\bqt}{\begin{quote}}
\newcommand{\eqt}{\end{quote}}
\newcommand{\bc}{\begin{center}}
\newcommand{\ec}{\end{center}}
\newcommand{\bdes}{\begin{description}}
\newcommand{\edes}{\end{description}}
\newcommand{\bfr}{\begin{flushright}}
\newcommand{\efr}{\end{flushright}}

\newcommand{\bpack}{\begin{list}{$\bullet$}{\parsep 0pt \itemsep 4pt \topsep 4pt \parskip 0pt \partopsep 0pt \leftmargin 28pt}}
\newcommand{\epack}{\end{list}}
\newcommand{\ben}{\bpack}
\newcommand{\een}{\epack}

\newcommand{\mynewsection}[1]{\vspace{.3cm}\bc{\large #1}\ec\vspace{.1cm}}

\newtheorem{hypo}{HYPOTHESIS}

\def\titleinfo{
\vspace{4mm}
\begin{center}
{\Large \bf Stressed and Unstressed Pronouns:\\
Complementary Preferences\\}
\vspace{15pt}
{\Large Megumi Kameyama\\}
\vspace{10pt}
{\large SRI International\\}
\end{center}
\vspace{15pt}
}

\begin{document}
%\twocolumn[\titleinfo]
\titleinfo

%\begin{document}
%\title{Stressed and Unstressed Pronouns:\\Complementary Preferences}
%\author{Megumi Kameyama\\SRI International}
%\date{}
%\maketitle

\noindent
{\bf Abstract}\\
\begin{footnotesize}
I present a unified account of interpretation preferences of stressed
and unstressed pronouns in discourse. The central intuition is the
Complementary Preference Hypothesis that predicts the interpretation
preference of a stressed pronoun from that of an unstressed pronoun
in the same discourse position. The base preference must be computed
in a {\em total} pragmatics module including commonsense preferences.
The focus constraint in Rooth's theory of semantic focus is
interpreted to be the salient subset of the domain in the local
attentional state in the discourse context independently motivated for
other purposes in  Centering Theory. 
\end{footnotesize}

\mynewsection{1 INTRODUCTION}

Stressed pronouns present a peculiar class of anaphoric expressions.
Informally, they communicate ``old'' as well as ``new'' information.
They are also at odds with the Topic--Focus Articulation (TFA)(Sgall,
Haji\v{c}ov\'{a}, and Panevov\'{a}, 1986) --- {\em contextually
bound}, which is typical of Topic, in spite of being in an
intonational focus, which is a defining property of Focus and common
for a {\em contextually unbound} expression.  They can also exemplify
the different notions of ``focus'' --- {\it psychological, semantic,}
or {\it contrastive} focus (Gundel, this volume) --- and can even
combine them all at once.  Unstressed pronouns (e.g., {\em he, she,
it, they}) and their stressed counterparts (e.g., {\em HE, SHE, THAT,
THEY}) often lead to drastically different interpretations.%
\footnote{Note that {\em it}
cannot be stressed, and its stressed counterpart is {\em THAT}.}
(Henceforth, stressed pronouns are in all uppercase.)

An account of the semantics and pragmatics of stressed pronouns
needs to explicate their peculiar hybrid behaviors and relation to
unstressed pronouns.
Example (1) below shows two alternative continuations of the first
utterance --- using an unstressed $he$ and stressed $HE$.  The
preferred interpretations indicated in the parentheses are opposite
from each other.
\enumsentence{\label{hit}
John hit Bill. Then \{$he\ \mid\ HE$\} was injured.\\ 
($he$ := Bill)($HE$ := John)
} Similarly, variations of Lakoff's (1971) example, (\ref{repub1}) 
and (\ref{repub2}) below,
show opposite preferred interpretations. 
\enumsentence{\label{repub1}
Paul called Jim a Republican. Then $he$ insulted $him$.\\ (Paul insulted Jim)
}
\enumsentence{\label{repub2}
Paul called Jim a Republican. Then $HE$ insulted $HIM$.\\ (Jim insulted
Paul)
} 
In (\ref{repub1}), the preferred interpretation of the unstressed
pronouns establishes grammatical parallelism between the two
utterances. In (\ref{repub2}), the preferred interpretation of the
stressed pronouns is the opposite from (\ref{repub1}), and the
discourse acquires an
additional assumption that ``x insults y'' follows from ``x calls y a
Republican.'' 

The present approach focuses on the interpretation differences between
the {\it stressed and unstressed counterparts} --- the stressed and
unstressed versions of the same pronominal form in the same {\it
position} in discourse as well as in a sentence.  I assume that
stressed and unstressed counterparts have exactly the same
denotational range --- the same range of {\it possible} values. They
also share the same interpretation problem of choosing the {\it
preferred} value.  The difference then comes from the latter.  What is
the difference in the preferred values of unstressed and stressed
counterparts? Is there a systematic relation between them?  Does one
preferred value {\it predict} the other?

I claim that the difference between stressed and unstressed
counterparts is in the {\em presuppositions}, and that there is a
systematic relation between them coming from the interaction between
the {\em semantic focus interpretation} of the stressed pronoun and
the {\em centering principles} associated with its unstressed
counterpart. The centering principles here are extended in the sense
that they are part of the {\em total} pragmatics that includes
commonsense preferences.  Under this unified approach, the systematic
relation between the stressed and unstressed counterparts is that of a
{\em complementary preference} within a suitable subset of the domain.

\mynewsection{2 THE STRESSED PRONOUN AND SEMANTIC FOCUS}

Rooth (1992) develops a theory of focus interpretation in terms of
{\em restricted alternative semantics}, where the focus semantics
introduces a presupposed constraint using a {\em focus interpretation
operator} $\sim$.  A focus phrase (or sentence) $\alpha$ has a focus
interpretation operator with a variable $C$ ($\sim C$). $C$ may denote
either individuals or sets of objects of the same type as the
denotation of $\alpha$, which are the elements or subsets of the {\em
focus semantic value} of $\alpha$ (henceforth $[[\alpha]]^f$). $C$
must contain the {\em ordinary semantic value} of $\alpha$ (henceforth
$[[\alpha]]^o$) and at least one more element or subset of $C$
distinct from $[[\alpha]]^o$, which is the {\em contrasting element or
subset} for $[[\alpha]]^o$.  Rooth points out that the value of $C$ is
a discourse entity to which $C$ is anaphoric, and that the important
question is how to constrain it.  This opens up focus interpretation
to all sorts of discourse--pragmatic effects. In this paper, I will
notationally distinguish two types of presuppositional constraints 
after Rooth (1993) ---
$C$ for a set of propositions and $F$ for a set of entities
(individuals or groups of individuals).

{\bf The Semantics of Stressed Pronouns.}  According to this theory of
semantic focus, a stressed $HE$ presupposes a constraint $\sim F$ that
there is a contextually determined set of entities ($[[HE]]^f =\{x\mid
x \in F\subseteq E\}$ where $E$ is the domain of individuals) with at
least two members --- the denotation of $HE$ ($[[HE]]^o$) and at least
one more contrasting individual. $F$ is the {\em contextually
available alternatives} or the {\em focus semantic value} of
HE. $[[HE]]^o$ is an element of $F$, and $F$ is a subset of $E$
($[[HE]]^o\in [[HE]]^f =F\subseteq E$).  An utterance (i.e., a
sentence token in a context) with a focused element presupposes a
constraint $\sim C$ that there is a contextually determined set of
propositions obtained by instantiating a set abstraction with the
alternative values of the focused element --- for instance, ``$HE$ was
injured'' presupposes a constraint $\sim C$ whose value is a set of
propositions obtained by instantiating the focus semantic value $\{
injured(x)\mid x\in E\}$ with the alternative values of
$[[HE]]^f$. The ordinary semantic value of the utterance is a truth
value that is an element of the focus semantic values ($[[HE\ was\
injured]]^o\in\{injured(x)\mid x\in F\}\subseteq\{injured(x)\mid x\in
E\}$).

{\bf The Pragmatics of Stressed Pronouns.}  Given the above focus
semantics, interpreting a stressed $HE$ in an utterance $U$ involves
four pragmatic subroutines. These subroutines are listed approximately
in the following bottom--up processing order, but no strict sequential
order is assumed: 
\ben
\item {\em Locate} $F\subseteq E$, where $F$ is 
a contextually determined set of entities that may contain more than
one potential referent for $HE$.
\item {\em Choose} a member of $F$ as $[[HE]]^o$.
\item {\em Discharge} $\sim C$ for $U$
(which may contain other focused phrases) in the current discourse
context --- in terms of a pending question or contrasting proposition.
\item {\em Establish Coherence} of $[[U]]^o$ in the current discourse
context; that is, contribution of the utterance content to the
evolving information state must somehow make sense.  
\een 
Each
pragmatic subroutine consists of an interacting set of preferences,
and the combination of all the subroutines 
may or may not converge into a single preferred interpretation of
$U$. We will focus here on the preferences that affect the {\em
locate} and {\em choose} subroutines, with illustrations of how they
interact with the {\em discharge} subroutine.  How is the relevant set
of alternatives located? How is the preferred value chosen? Rooth does
not discuss these cases where focus interpretation may also involve
choosing among multiple alternatives. This paper proposes an approach
that spells out the missing detail in 
his notion of restricted alternatives.

\mynewsection{3 THREE BACKGROUND HYPOTHESES}

Our present aim is to account for the systematic interpretation
difference between the stressed and unstressed counterparts. I
motivate here the claim that the interpretation of the {\it
unstressed} counterpart should be the basis from which to predict the
interpretation of its stressed counterpart. The claim rests on the
following analogous characterization of the semantics and pragmatics
of the unstressed pronoun such as {\it he} and {\it him}.

An unstressed {\it he} presupposes a constraint $\sim B$ that there is
a contextually determined set of entities ($[[he]]^{f'} =\{x\mid x \in
B\subseteq E\}$) with at least {\em one} member --- the denotation of
{\it he} ($[[he]]^o$). $[[he]]^o$ is an element of $B$ and $B$ is a
subset of $E$ ($[[he]]^o\in [[he]]^{f'} =B\subseteq E$).  Given this
semantics, interpreting an unstressed {\it he} in an utterance $U$
involves the following pragmatic subroutines:
\ben
\item {\em Locate} $B\subseteq E$, where $B$ is a
contextually determined set of entities that may contain more than one
potential referent for $he$.
\item {\em Choose} a member of $B$ as $[[he]]^o$.
\item {\em Establish Coherence} of $[[U]]^o$ in the current discourse
context.
\een

Note that the sets of pragmatic subroutines for the stressed and
unstressed pronouns correspond one--on--one {\em except} for the
additional {\em discharge} constraint on the utterance containing a
stressed pronoun. This indicates that the sum of the pragmatic
constraints on the use of stressed pronouns may be greater than on
their unstressed counterparts. I therefore hypothesize the following:
\begin{hypo}
Given the range $\beta$ of felicitous uses of {\em unstressed}
pronouns in discourse and the range $\alpha$ of felicitous uses of
their {\em stressed} counterparts, $\alpha\subset\beta$.
\end{hypo}
The above hypothesis is borne out by examples of {\em asymmetry}
such as the following:
\enumsentence{\label{babar}
Babar went to a bakery.  $he$ greeted the baker.\\  \{$he\ \mid\ ??HE$\}
pointed at a blueberry pie.  } The infelicity of the stressed $HE$ is
due to the difficulty in discharging the presupposed focus constraint
on the utterance (``x pointed at a blueberry pie'') in terms of a
pending question (``Who pointed at a blueberry pie?'') or contrasting
proposition (e.g., ``someone did not point at a blueberry pie'').

Assuming the above hypothesis, we may conclude that if the preferred
interpretation of either the stressed or unstressed counterpart
serves as the basis for deriving the other preferred interpretation,
the base preference should come from the unstressed counterpart, hence
the second background hypothesis:
\begin{hypo}
The preferred value of a stressed pronoun can be predicted by the
preferred value of its unstressed counterpart.
\end{hypo}

There are remaining questions. What is the ``contextually relevant
subset'' $B$ for the unstressed pronoun?  What constraint, if any,
does an utterance containing an unstressed pronoun presuppose? In
other words, is there an unstressed pronoun analogue of $\sim C$? I
will propose an answer only to the first of these questions.  I will
define the {\em currently salient subset} of the domain of individuals
in terms of the centering model of discourse (see the next section),
and motivate the third background hypothesis:
\begin{hypo}
Stressed and unstressed counterparts choose their values from the {\em
same salient subset} of the domain of individuals (i.e., $F=B$).
\end{hypo}
If we are lucky, then, what we have is an integrated account of the
semantics and pragmatics of stressed and unstressed pronouns. Note
that in order for this hypothesis to work, we need to explain the
discrepancy that the set $F$ for a stressed pronoun must contain at
least two members, whereas $B$ may sometimes contain only one. I will
argue that when the salient subset is a singleton, the contrasting
members are accommodated into the context to satisfy the focus
constraint of the stressed pronoun.

\mynewsection{4 THE UNSTRESSED PRONOUN AND CENTERING}

The problem of choosing among alternative values for pronouns has been
investigated in the framework of Centering Theory (Grosz, Joshi,
and Weinstein, 1983, 1986, 1995).\footnote{Centering Theory 
synthesizes previous work on discourse
focusing (Grosz, 1977, 1981), immediate focusing (Sidner, 1979, 1983), and
discourse centering (Joshi and Kuhn, 1979; Joshi and Weinstein,
1981).}  It is part of an overall theory of discourse structure and
meaning (Grosz and Sidner, 1986) that distinguishes among three
components of discourse structure --- a linguistic structure, an
intentional structure, and an attentional state --- and two levels of
discourse coherence --- global and local. {\em Attentional state}
models the discourse participants' focus of attention determined by
the intentional and linguistic structures at any given point in the
discourse. It has global and local components corresponding to the two
levels of discourse coherence. The global--level component is a stack
of focus spaces, where each focus space holds entities and
propositions associated with a discourse segment, which is associated
with a discourse segment purpose in the intentional structure.
Centering models the local--level component of attentional state  ---
how the speaker's linguistic choices for describing propositional
contents affect the {\em inference load placed upon the hearer} in
discourse processing.

An utterance in discourse (not a sentence in isolation) has entities
called {\em centers} that link the utterance with other utterances in
the same discourse segment. They are the set of {\em forward--looking
centers} (Cf) partially ordered by relative prominence. One member of
the Cf may be the {\em backward--looking center} (Cb) that connects
with a member of the Cf of the previous utterance. The speaker's
linguistic choices define centering transitions that affect the local
coherence of the discourse.  In English discourse, pronouns and
grammatical subjects are the main indicators of centering transitions
(Grosz et al., 1983, 1986, 1995; Kameyama, 1985, 1986; Brennan,
Friedman, and Pollard, 1987). {\em Unstressed} pronouns, in
particular, are primarily used to indicate the Cb in English--type
languages (Kameyama, 1985:Ch.1).

{\bf Dynamic Preference Model.}  
Kameyama (1996) develops an initial model of {\em interacting
preferences} for dynamically updating a multicomponent context data
structure in discourse processing.  Centering preferences, stated in
terms of the attentional notion of {\em salience}, systematically
interact with structural and commonsense preferences to predict the
preferred interpretation of unstressed pronouns.  
The perspective is that of {\it total} pragmatics that includes both
linguistic and commonsense preferences.
This {\em dynamic
preference model} is summarized below. (A preference is stated
as a defeasible rule in the form of either ``normally $p$'' or ``if
$p$ then normally $q$.'')
\bit
\item {\em Discourse} is a sequence of utterances,
$U_1,\ldots,U_n$.  Each utterance $U_i$ defines a {\em
transition relation} between the {\em input context} $C_{i-1}$ and the
{\em output context} $C_i$.\footnote{Utterance $U$ need not be 
a sentence in the standard syntactic sense. In fact, 
it is more natural to
think of $U$ as a tensed clause, allowing multiple intrasentential
transitions within a complex sentence (see Kameyama, to appear).
This paper focuses only on intersentential anaphora, however.}

\item Context $C$ is a multicomponent data
structure $C_i =\langle\phi_i^k,A_i,D_i,\ldots\rangle$
including:\footnote{Indexical Context is notably omitted here.}
\bit
\item  $\phi_i^k$ (LF Register) --- the preferred interpretation 
$k$ of the last utterance $U_i$ in a logical
form that preserves aspects of the syntactic structure of $U_i$.

\item $A_i$ (Attentional State) --- a set of currently ``open'' propositions
with the associated entities, into which a new utterance content can
potentially be integrated. The entities in $A_i$ are partially ordered
by {\it salience}, and the most salient subpart of $A_i$ is the {\em
local attentional state}, $A_i^{LOC}$ (see below).

\item $D_i$ (Discourse Model) --- a structured information state for 
what the discourse has been about (situations, eventualities,
entities, and relations among them), including the content of $A_i$.
\eit

\item $A_i^{LOC}$ (Local Attentional State)\footnote{There are other highly salient entities
that we are not considering in this paper --- for instance, the
entities associated with the current discourse segment {\it purpose},
the old $Center$ entities in the current discourse segment, and the
entities realized by possible antecedents in the current utterance. }\label{aloc}
 ---  
the entities realized\footnote{In the examples in this paper, {\em
realize} simply means {\em denote}, but it combines a variety of
factors in a full account (see Grosz et al., 1995).} by $\phi_i^k$
(corresponding to the Cf). One of them may be $Center_i$
(corresponding to the Cb)\footnote{$Center$ corresponds to ``topic
proper'' in TFA (Sgall et al., 1986).} --- the entity on which
the current discourse segment is centered. I assume that 
$Center_i$ may
be missing in $A_i$, especially at the onset of a new discourse
segment. 

\item In utterance interpretation, 
there are interacting preferences for updating different context
components. Some come from the linguistic knowledge, and others come
from the world knowledge.
\eit

{\bf The Role of the Attentional State.}  
Various factors affect salience dynamics
--- including utterance forms, discourse participants' purposes and
perspectives, and the perceptually salient objects in the utterance
situation. Here we focus on the factor of utterance forms.

Two default linguistic
hierarchies are relevant to the dynamics of salience --- {\it
grammatical function hierarchy} (GF ORDER) and the {\it nominal
expression type hierarchy} (EXP ORDER):\footnote{Both linguistic hierarchies are
in fact recurrent in functional and typological studies of language.
The GF ORDER closely resembles Keenan and Comrie's (1977)
Accessibility Hierarchy, Givon's (1979) Topicality Hierarchy, and
Kuno's (1987) Thematic Hierarchy, all of which predict the preferred
syntactic structure for describing the things that a sentence is
``mainly about'' within and across languages.  The EXP ORDER resembles
the linguistic correlates of Gundel, Hedberg, and Zacharski's (1993)
Givenness Hierarchy, which is closely related to Prince's (1981)
Familiarity Scale, which predicts the relative {\it degrees} of
accessibility of referents. It is of interest that virtually the same
hierarchies are relevant to the computational interest in how grammar
controls inferences in language use.}  
\bdes
\item[GF ORDER:] Given a hierarchy [{\sc SUBJECT $>$ OBJECT $>$ OBJECT2 $>$
Others}], an entity realized by a higher--ranked phrase is
normally more salient in the {\em output} attentional state.

\item[EXP ORDER:] Given a hierarchy [{\sc Zero
Pronominal $>$ Pronoun $>$ Definite NP
$>$ Indefinite NP}],%
\footnote{The distinction between stressed and
unstressed pronouns in EXP ORDER in an earlier version (Kameyama,
1994) has been removed because the ordering {\sc Unstressed Pronoun
$>$ Stressed Pronoun} follows from the present 
independent account of stress.}
 an entity realized by a higher--ranked expression
type is normally more salient in the {\em input} attentional state.
\edes 
Since matrix subjects and objects cannot be omitted in English, the
highest--ranked expression type is the (unstressed) pronoun (Kameyama,
1985:Ch.1). From EXP ORDER, it follows that a pronoun
normally realizes a {\it maximally salient entity} (of an appropriate
gender--number--person type) in the input $A$. This
accounts for the preference for a pronoun to corefer with the matrix
subject in the previous utterance as in the following
example:\footnote{I will henceforth notate the relation
$more\_salient\_than$ with $>$, indeterminate salience ordering with
$<>$, $preferred\_over$ with $\prec$, $weakly\_preferred\_over$
with $\prec_?$, and indeterminate preference ordering with $\prec\succ$. Preferences in the subsequent examples come from the
survey data discussed in Kameyama (1996).}
\eenumsentence{\label{home}
\item[1.] John hit Bill. 
\hspace{3.5cm}$A_1$:[[John$>$Bill]$_{\phi_1}$\ldots]
\item[2.] Mary told him to go home. 
\hspace{1.5cm}$him$ := John$\prec$Bill
}
The centering model is reinterpreted as follows:
\bdes
\item[CENTER:] $Center_i\in A_i$ is normally more salient than other
entities in $A_i$.
\item[EXP CENTER:] An expression of the highest--ranked type 
in $U_i$ normally realizes $Center_i$ in the {\it output}
attentional state $A_i$.\footnote{The ``highest--ranked type'' in
EXP CENTER can be interpreted as either relative to each utterance or
absolute in all utterances. Under the relative interpretation, a
nonpronominal expression type can also output the Center as long as
there are no pronouns in the same utterance.  Under the absolute
interpretation, only the pronominals (either zero or overt, depending
on the syntactic type of the language) can output the Center.  I will
take the absolute interpretation in this paper following Kameyama
(1985, 1986), based on the rationale that the choice of the
highest--ranked pronominal forms in a language should reflect a
certain absolute sense of salience threshold.}
\edes
EXP ORDER and EXP CENTER combine to make a pronoun either {\it
establish} or {\it chain} the Center (Kameyama, 1985,
1986).\footnote{{\em Chain} corresponds to what I have previously
called {\it retain}. It covers both CONTINUE and RETAIN distinguished
in Grosz et al. (1986, 1995).}  
$Center_i$ is ``established'' ($Center_{i-1}\neq Center_i$) when a
pronoun picks a salient non--Center in $A_{i-1}$ and makes it the
Center in $A_i$.  It is ``chained'' ($Center_{i-1}=Center_i$) when a
pronoun picks $Center_{i-1}$ and outputs it as $Center_i$.

The {\em maximally salient entity} may be determinate in some
attentional state and indeterminate in others, depending on whether
the GF--based salience ordering and the Center--based one converge.  A
highest--ranked GF and the Center may converge or diverge in the {\it
input} $A$, affecting the preferred interpretation of a
pronoun.\footnote{Discussions with Becky Passonneau helped clarify
this point.} For instance, the Center realized by a sentence--initial
matrix subject pronoun is the single most salient entity, but the
Center realized by a nonsubject pronoun competes with the entity
realized by the subject, resulting in an indeterminacy.\footnote{Under
the present perspective, the
proposal to distinguish between CONTINUE and RETAIN transitions (Grosz
et al., 1986, 1995) focuses on the convergence 
of the highest--ranked GF
(Cp) and the Center (Cb) in the {\it output} attentional state.}
The different effects are illustrated here:
\eenumsentence{\label{babar1}
\item[1.] Babar went to a bakery.
\hspace{2cm}$A_1$:[[Babar$>$Bakery]$_{\phi_1}$\ldots]
\item[2.] $he$ greeted the baker.
\\\hspace{3cm}$A_2$:[[[Babar]$^{Subj}_{Center}$$>$Baker]$_{\phi_2}$$>$Bakery\ldots]
\item[3.] $he$ pointed at a blueberry pie.
\hspace{1.5cm}$he$ := Babar$\prec$Baker
} 
\eenumsentence{\label{babar2}
\item[1.] Babar went to a bakery.
\hspace{2cm}$A_1$:[[Babar$>$Bakery]$_{\phi_1}$\ldots]
\item[2.] The baker greeted $him$.
\\\hspace{3cm}$A_2$:[[Baker$<>$[Babar]$^{Obj}_{Center}$]$_{\phi_2}$$>$Bakery\ldots]
\item[3.] $he$ pointed at a blueberry pie.
\hspace{1.5cm}$he$ := Baker$\prec_?$Babar
} 
Example (\ref{babar1}) shows the effects of determinate salience
ranking in terms of a chain of subject Centers. 
The preferred value of $he$ in (\ref{babar1})--3 is determinate.
In contrast, the
salience ranking in $A_2$ is indeterminate in example (\ref{babar2}).
The weak preference in (\ref{babar2})--3 comes from the interaction
of attentional preference and the separate preference for grammatical
parallelism stated below.

{\bf The Role of the LF Register.}  The grammatical parallelism of two
adjacent utterances in discourse affects the preferred interpretation
of pronouns (Kameyama, 1986), tense (Kameyama, Passonneau, and Poesio,
1993), and ellipses (Pruest, 1992; Kehler, 1993). This general
tendency warrants a separate statement.  Parallelism is achieved, in
the present account, by a computation on the pair of logical forms,
one in the LF register in the context, and the other being
interpreted: \bdes
\item[PARA:] The LF register in the input context and the utterance
being interpreted seek maximal parallelism.\footnote{For a specific
definition of parallelism, see, e.g., Pruest (1992), Kameyama (1986).}
\edes
We have observed in example (\ref{babar2}) that this parallelism
preference kicks in when the salience--based preference is
indeterminate. 

{\bf The Role of the Discourse Model.}  Both linguistic semantics and
commonsense preferences apply on the same discourse model. Lexically
triggered conventional presuppositions, for instance, constrain
possible discourse models. Preferential rules assign a partial order
on these models. Commonsense preferences consist of all that an
ordinary speaker knows about the world and life. There will be a
relatively small number of linguistic pragmatic rules that
systematically interact with and {\em control} an open--ended mass of
commonsense rules.  Our aim is to describe the former as fully as
possible, and specify how the ``control mechanism'' works. Linguistic
rules should be stable across examples and domains, while there will be
different commonsense rules for each new example and domain. Example
(\ref{hit}) illustrates a type of causal knowledge:
\bdes
\item[HIT:] When an agent x hits an agent y, y is normally hurt.
\edes

{\bf Preference Interactions.}  Preferences relevant to unstressed
pronoun interpretation fall into three {\em preference classes}
corresponding to the preferred transitions of the three context
components.  CENTER, GF ORDER, EXP ORDER, and EXP CENTER are
defeasible {\it Attentional Rules} (ATT) stating the preferred
$A$--transitions.  PARA is an example of defeasible {\it LF Rules}
(LF) stating the preferred LF--transitions. HIT is an example of
defeasible {\it Commonsense Rules} (WK) stating the preferred
$D$--transitions. These preference classes independently conclude the
preferred interpretation of an utterance, and these class--internal
conclusions combine in a certain general 
pattern to produce the final preference. 
Crucially, preferences {\it can
override} other preferences that contradict them. Ambiguities persist
only when mutually contradictory preferences are equally strong.

We have identified the 
following general patterns of preference interactions in pronoun
interpretation:\footnote{See Jaspars and Kameyama (1996) for a model
of discourse logic that incorporates the notion of preference classes.}
\bpack
\item Indefeasible syntax and semantics (SYN+SEM) can override all
preferences.
\item Commonsense preferences can override attentional or parallelism
preferences.  This overriding can be difficult, however,
 when the latter is
extremely strong, producing garden--path phenomena.  Observe the
difference between ``{\it John hit Bill. He was severely injured.}''
($he$ := Bill$\prec$John) and ``{\it Tommy came into the classroom. He
saw Billy at the door. He hit him on the chin. ??He was severely
injured.}'' ($he$ := Tommy is first chosen but retracted, and results
in Billy).
\item Attentional preferences can override 
parallelism preferences except
for the cases of parallelim induced by conventional presuppositions.
\epack

This general overriding pattern is schematically shown
here, where $\geq$ represents a ``can override'' relation:
\vspace*{2mm}
\begin{center}
\begin{tabular}{|ccccccc|}\hline
SYN+SEM & $\geq$ & WK & $\geq$ & ATT & $\geq$ & LF
\\\hline
\end{tabular}
\end{center}\ \\

We have thus a fairly rich theory of interacting preferences in
unstressed pronoun interpretation.  The question is the following ---
are they related to, or better, predictive of the preferences relevant
to the stressed counterpart?

\mynewsection{5 COMPLEMENTARY PREFERENCE HYPOTHESIS}

I claim that the preferred value of a stressed pronoun in discourse is
predictable from the preferred value of its unstressed counterpart,
and that they draw their values from the same `currently salient'
subset of the domain.  This salient subset is the presupposed
constraint $F$ for the stressed pronoun and $B$ for the unstressed
pronoun in Hypothesis 3. 

In this unified view, this salient subset is the {\em local
attentional state}, $A^{LOC}$, or the ``center of attention'' in the
dynamic context. 
\bqt
{\bf Salient Subset Hypothesis (SSH):} 
{\em An
unstressed pronoun and its stressed counterpart 
in utterance $U_i$ draw their possible values from
the {\it input} local attentional state} ($A_{i-1}^{LOC}=F=B$).
\eqt
Although the exact specification of $A^{LOC}$ is still open
(see note 5), we take the standard view in Centering
Theory as the starting point 
in this paper --- it is the set of entities realized by the
previous utterance in the given discourse segment. It corresponds to
the entities associated with the logical form in the LF register in the
present dynamic preference model.

Given the common
presupposed subset of the domain, I hypothesize that the preference
order among the alternative values for a stressed pronoun is the {\em
complement} of the preference order for its unstressed counterpart:
\bqt 
{\bf Complementary Preference Hypothesis (CPH):} {\em A focused
pronoun takes the complementary preference of the unstressed
counterpart.}
\eqt 
Restricting the salient subset is also crucial in this
account, in order to avoid an infinite regression into the least
preferred entity in the entire universe of discourse to get at the
most preferred value for a stressed pronoun. 

{\bf Computation of the Preferred Value.}  Given the CPH, the
preferred value of a stressed pronoun in utterance $U_i$ is computed
with the following algorithm:\footnote{The subset $H_{i-1}$ was 
overlooked in
the earlier formulation of the algorithm in Kameyama (1994).}
\ben
\item {\em Locate} the local attentional state $B_{i-1}$ in the input
context $C_{i-1}$. $B_{i-1}$ contains a nonempty set of entities
partially ordered by {\em salience}
($B_{i-1}^{salience}$).\footnote{Actually, this set can also be empty,
in which case a plausible assumption is that
 an accommodation takes place and obtains an indexically
salient entity.}
\item {\em Compute Base Preference Order} for $B_{i-1}^{salience}$ for 
the {\em unstressed counterpart} of the stressed pronoun, in terms of
the interaction of LF rules, attentional rules, and commonsense
rules.\footnote{The salience order controls this computation.}  
The output is a subset $H_{i-1}$ of $B_{i-1}$ partially ordered by {\em
preference} ($H_{i-1}^{pref}$), where $H_{i-1}\subseteq B_{i-1}$
contains only the possible values of the pronoun within $B_{i-1}$.
\item {\em Compute Complementary Preference Order} for
$H_{i-1}^{pref}$ as follows: x$\prec$y becomes y$\prec$x, and
x$\prec\succ$y does not change.  
With the CPH, this outputs the possible
values of the stressed pronoun partially ordered by preference
($H_{i-1}^{pref^\cup}$).
\item {\em Discharge} the presupposed constraint $\sim C$ for $U_i$
(which may contain other focused phrases).
If $B_{i-1}$ is a singleton set, at least one additional
contrasting individual 
is {\em accommodated} (see below). 
\item {\em Establish Coherence} of $[[U_i]]^o$.
\een
The algorithm is illustrated with example (\ref{hit}) (repeated here):\\
\ben
\item[1.] John hit Bill.
\hspace{2cm}$C_1$: $B_1^{salience}$=\{John$>$Bill\}
\item[2b.] Then $HE$ was injured.\\\hspace{2cm} 
$B_1\supseteq H_1^{pref}$=\{Bill$\prec$John\} $\leadsto$
$H_1^{pref^\cup}$=\{John$\prec$Bill\}\\\hspace{2cm} 
$\leadsto$ $HE$ := \{John$\prec$Bill\} 
\een 
The local
attentional state in the input context for $U_{2b}$ contains John and
Bill, with John more salient than Bill.  Both John and Bill are
possible values of $he$.  The base preference for the unstressed
counterpart results from the WK rule (HIT) overriding ATT and LF
rules, with Bill preferred over John. The complementary preference
makes John preferred over Bill for the stressed pronoun. This
preference then survives the presupposition discharging, which
recognizes the (indirect) contrast between ``John was injured'' and
``John hit Bill'' and coherence establishing, which recognizes the
Cause--Effect relation between $U_1$ and $U_{2b}$.

The following variant of (\ref{hit}) illustrates a case where not all
the entities in $B$ are possible values for the pronoun:
\ben
\item[1.] John hit Bill in front of Mary.\\
\hspace{2cm}$C_1$: $B_1^{salience}$=\{John$>$Bill$>$Mary\}
\item[2.] Then $HE$ was injured.\\\hspace{2cm} 
$B_1\supseteq H_1^{pref}$=\{Bill$\prec$John\} $\leadsto$
$H_1^{pref^\cup}$=\{John$\prec$Bill\}\\\hspace{2cm} 
$\leadsto$ $HE$ := \{John$\prec$Bill\} 
\een 
Here, $B_1$ contains three individuals, John, Bill, and Mary, of
which only two are possible values of $he$. The interpretation of 
stressed $HE$ in $U_{2}$ is exactly the same as in the previous
example, however, since the base preference is on the same subset $H_1$. 

Nakatani (1993) found in spontaneous narratives that a stressed
subject pronoun tends to signal (1) a {\em local shift} to a non--Cb
in the Cf or (2) a {\em global shift} to an old Cb within the segment.
(1) is highly consistent with the present account. (2) would also
follow if the local attentional state contains old Cbs in the segment
as well as the Cf.

%There is a question about exactly how many entities can belong to $H$,
%the set on which the preference complementation applies. If the above
%example starts out with ``John hit Bill in front of George,'' for
%instance, is George the preferred value of a stressed $HE$? This
%question is related to whether George is within the range of entities
%that can be referred to with an unstressed counterpart $he$. These
%issues must be resolved in the future.

{\bf Indeterminate Preferences.} This unified account predicts that
when the preferred value of an unstressed pronoun is indeterminate,
the preferred value for its stressed counterpart is likewise
indeterminate because the complement of an indeterminate order is
still indeterminate. This prediction is borne out in the following
example:
\enumsentence{
Jack and Bob are good friends.  \{??$he\ \mid$ ??$HE$\} is from
Louisiana.  } Neither Jack nor Bob is more salient than the other in
the conjoined NP, and neither parallelism nor commonsense preferences
distinguish them,
hence the indeterminate preference for $he$. Its
complement for $HE$ is also indeterminate.

{\bf Unambiguous Stressed Pronouns.} The proposed mechanism
correctly predicts that when the pronoun in question is unambiguous,
stressing does not change its values. Since the stressed and
unstressed counterparts draw their possible values from the same
singleton set, $\mid H\mid =1$, the complementation operation
produces the same possible value for the stressed counterpart. 
The following examples illustrate such unambiguous pronouns:
\enumsentence{\label{jack2}
Jack and Mary are good friends.
\{$he\ \mid\ HE$\} is from Louisiana.\\
\{$he\ \mid\ HE$\} := {Jack}
} 
\enumsentence{\label{jack1}
Jack is a physicist.
\{$he\ \mid\ HE$\} is from Louisiana.
\\\{$he\ \mid\ HE$\} := {Jack}
}
The interpretation process of the 
stressed $HE$ in example (\ref{jack2}) is illustrated here:
\eenumsentence[\ref{jack2}]{
\item[1.] Jack and Mary are good friends.\\
\hspace{2cm}$C_1$: $B_1^{salience}$=\{Jack$<>$Mary\}
\item[2b.] $HE$ is from Louisiana.\\\hspace{2cm} 
$B_1\supseteq H_1^{pref}$=\{Jack\} $\leadsto$
$H_1^{pref^\cup}$=\{Jack\} $\leadsto$ $HE$ := \{Jack\} 
}
The preference complementation operation yields the correct
unambiguous value of
the stressed pronoun, $[[HE]]^o$, 
with no additional stipulations. A possible  
alternative to the CPH such as ``rule out the most preferred value
in $H_1^{pref}$'' would not
naturally extend to an account of unambiguous pronouns.

Recall the focus constraint $\sim F$ on stressed pronouns. The focus
semantic value of $HE$, $F$, must have at least two members --- the
denotation of $HE$ ($[[HE]]^o$) and at least one more contrasting
individual. 
These contrasting individuals also instantiate contrasting
propositions that discharge the focus constraint $\sim C$ on the
utterance as a whole.
Under the present proposal, $F$ is the set of
entities in the local attentional state, which is
 also presupposed for
unstressed pronouns ($F=B=A^{LOC}$), so 
$A^{LOC}$ must provide the contrasting individuals.
The question is --- are they always supplied from $A^{LOC}$ rather
than $A$? 
 
The present account makes use of nested subsets of the entities in $A$
--- $H\subseteq B=A^{LOC}\subseteq A$. 
Our examples in this paper indicate that these regions are all
potential sources for the contrasting individuals. We have seen the
following three sources for contrasting individuals:
\ben
\item Alternative values within $H\subseteq A^{LOC}$ --- 
examples (\ref{hit}) and (\ref{repub2})
\item Individuals in $A^{LOC}$
that are not possible values of the pronoun but belong to a common
general class with the pronoun referent 
such as {\sc person} --- example (\ref{jack2})%
\footnote{I assume a general sort hierarchy for the domain of
individuals as proposed by, e.g., Bosch (1988) and Prevost (1996), to
compute the notion of `a common general class' under which a contrast
is established.}
\item Not found in  $A^{LOC}$ but potentially found in $A$ 
--- example (\ref{jack1}). 
\een
I will discuss what may happen in the third
case.

When $A^{LOC}$ does not provide contrasting individuals, the focus
constraint for the stressed pronoun $\sim F$ is satisfied by
discharging the focus constraint on the utterance $\sim C$ by {\em
accommodation} (Lewis, 1979) into the input attentional state. For
example, utterance $U_{2b}$ in both (\ref{jack2}) and (\ref{jack1})
generates the same focus constraint $\sim C$ --- ``x is from
Louisiana.''  In (\ref{jack2}), $A^{LOC}$ contains the contrasting
individual, Mary, and $\sim C$ is discharged by accommodating a
contrasting presupposition ``Mary is not from Louisiana.'' In
(\ref{jack1}), $A^{LOC}$ does not contain contrasting individuals, and
$\sim C$ is discharged by accommodating a question ``Who is from
Louisiana?''  This question presupposes other persons in the domain,
and if no persons have been explicitly mentioned, a set of persons is
implicitly accommodated into the current attentional state $A$.
A more precise formulation of this process will be left for a future
task. 

The present framework offers an explicit proposal about how the
focus constraints at both the phrase and utterance levels are
discharged within the nested structure of attentional state, and how
contrasts are accommodated when presuppositions do not immediately
follow from the explicit discourse.

\mynewsection{6 FURTHER QUESTIONS}

There are a number of related questions, which I will only note
here. 
\bit
\item The local attentional state is supposed to be relevant only for
utterance processing within a discourse segment. What happens to
stressed and unstressed pronouns in a segment--initial utterance? More
generally, how does the {\em discourse structure} affect pronoun
interpretation?  
\item {\em Intrasentential} pronominal anaphora for both
stressed and unstressed pronouns is widely studied in syntax
(e.g., Akmajian and Jackendoff, 1970; Lakoff, 1976;
Williams, 1980; Lujan, 1985; Hirschberg and Ward, 1991).  
How does the present account relate to these syntactic facts?  
How
does it extend to the pragmatics of intrasentential pronominal
anaphora (e.g., Kameyama, to appear)?  
\eit

\mynewsection{7 CONCLUSION}

I have presented a unified account of interpretation preferences of
stressed and unstressed pronouns in discourse. The central intuition
is expressed as the Complementary Preference Hypothesis taking the
interpretation preference of the unstressed pronoun as the base from
which to predict the interpretation preference of the stressed pronoun
in the same discourse position. This base preference must be computed
in a {\em total} pragmatics module including commonsense
preferences. I have also made a concrete proposal for the
pragmatically determined focus constraint in Rooth's theory of
semantic focus. The salient subset of the domain in this proposal
makes use of the dynamically updated local attentional state in the
discourse context independently motivated for other purposes in
Centering Theory. As a consequence, the overall discourse processing
can unify the interpretation process for the two kinds of pronouns
while explaining the source of the curious complementarity in their
interpretation preferences.

\mynewsection{NOTES}
\begin{footnotesize}
% This was for endnotes.sty from CSLI
%\listofendnotes 
% This is for CTAN endnotes.sty
\begingroup
\parindent 0pt
\parskip 2ex
\def\enotesize{\normalsize}
\theendnotes
\endgroup

\noindent
{\bf Acknowledgments}\\
This work was in part supported by the National Science
Foundation and the Advanced Research Projects Agency under Grant
IRI--9314961 (Integrated Techniques for Generation and Interpretation).
I would like to thank Peter Bosch, Elisabet Engdahl, and Hiroshi
Nakagawa for helpful comments on an earlier version.

\mynewsection{REFERENCES}

\noindent
\begingroup
\parindent=-2em \advance\leftskip by2em
\parskip=5pt

Akmajian, Adrian and Ray Jackendoff. 1970. Coreferentiality and
Stress. {\it Linguistic Inquiry}, 1(1), 124--126.

Bosch, Peter. 1988. Representing and Accessing Focussed
Referents. {\it Language and Cognitive Processes}, 3(3), 207--231.

Brennan, Susan, Lyn Friedman, and Carl Pollard. 1987.  A Centering
Approach to Pronouns.  In {\it Proceedings of the 25th Annual Meeting
of the Association for Computational Linguistics}, 155--162.

Givon, Talmy. 1979. {\it On Understanding Grammar}, Academic Press,
New York, NY.

Grosz, Barbara. 1977. The Representation and Use of Focus in Dialogue
Understanding. Technical Report 151, SRI International, Menlo Park,
CA.

Grosz, Barbara. 1981. Focusing and Description in Natual Language
Dialogues. In Joshi, Aravind, Bonnie Webber, and Ivan Sag, eds. {\it
Elements of Discourse Understanding}, Cambridge University Press,
Cambridge, England, 85--105.

Grosz, Barbara, Aravind Joshi, and Scott Weinstein. 1983.  Providing a
Unified Account of Definite Noun Phrases in Discourse.  In {\it
Proceedings of the 21st Meeting of the Association of Computational
Linguistics}, 44--50.

Grosz, Barbara, Aravind Joshi, and Scott Weinstein. 1986.  Towards a
Computational Theory of Discourse Interpretation.  Unpublished
manuscript. [The final version appeared as Grosz et al. 1995.]

Grosz, Barbara, Aravind Joshi, and Scott Weinstein. 1995. 
Centering: A Framework for Modelling
the Local Coherence of Discourse. In {\it Computational
Linguistics}, 21(2), 203--226.

Grosz, Barbara and Candy Sidner. 1986. Attention, Intention, and the
Structure of Discourse. {\it Computational Linguistics}, 12(3),
175--204.

Gundel, Jeanette, Nancy Hedberg, and Ron Zacharski. 1993. Cognitive
Status and the Form of Referring Expressions in Discourse. {\it
Language}, 69(2), 274--307.

Hirschberg, Julia and Gregory Ward. 1991. Accent and Bound
Anaphora. {\it Cognitive Linguistics}, 2(2), 101--121.

Jaspars, Jan and Megumi Kameyama. 1996.  Preferences in Dynamic
Semantics.  In Paul Dekker and Martin Stokhof, eds., {\it Proceedings
of the 10th Amsterdam Colloquium}, Institute for Logic, Language, and
Computation, University of Amsterdam, Amsterdam, 445--464.

Joshi, Aravind and Steve Kuhn. 1979. Centered Logic: The Role of
Entity Centered Sentence Representation in Natural Language
Inferencing. In {\it Proceedings of International Joint Conference on
Artificial Intelligence}, Tokyo, Japan.

Joshi, Aravind and Scott Weinstein. 1981. Control of Inference: Role
of Some Aspects of Discourse Structure --- Centering. In {\it
Proceedings of International Joint Conference on Artificial
Intelligence}, Vancouver, Canada, 385--387.

Kameyama, Megumi. 1985. {\it Zero Anaphora: The Case of Japanese.}
Ph.D. Thesis, Stanford University.

Kameyama, Megumi. 1986. A Property-sharing Constraint in Centering.
In {\it Proceedings of the 24th Annual Meeting of the Association for
Computational Linguistics}, New York, NY, 200--206.

Kameyama, Megumi. 1994. 
Stressed and Unstressed Pronouns: Complementary Preferences. In Bosch,
Peter and Rob van der Sandt, eds., {\it Focus and Natural Language
Processing}. Institute for Logic and Linguistics, IBM, Heidelberg,
475--484.

Kameyama, Megumi. 1996. Indefeasible Semantics and Defeasible
Pragmatics. In Kanazawa, Makoto, Christopher Pi\~{n}on, and
Henri\"{e}tte de Swart, eds., {\it Quantifiers, Deduction, and
Context}.  CSLI, Stanford, CA, 111--138.

Kameyama, Megumi. To appear. Intrasentential Centering. In Walker,
Marilyn, Ellen Prince, and Aravind Joshi, eds., 1997. {\it Centering
in Discourse}, Oxford University Press, Oxford, England.

Kameyama, Megumi, Rebecca Passonneau, and Massimo
Poesio. 1993. Temporal Centering. In {\it Proceedings of the 31st
Meeting of the Association of Computational Linguistics}, Columbus,
OH, 70--77.

Keenan, Edward and Bernard Comrie. 1977. Noun Phrase Accessibility
and Universal Grammar. {\it Linguistic Inquiry}, 8(1), 63--100.

Kehler, Andrew. 1993. The Effect of Establishing Coherence in Ellipsis
and Anaphora Resolution.
In {\it Proceedings of the 31st Meeting of the
Association of Computational Linguistics}, Columbus, OH, 62--69.

Kuno, Susumu. 1987. {\it Functional Syntax}, Chicago University Press,
Chicago, IL.

Lakoff, George. 1971. Presupposition and Relative Well--formedness. In
Steinberg, Danny and Leon Jakobovits, eds., {\it Semantics: An
Interdisciplinary Reader in Philosophy, Linguistics, and Psychology.}
Cambridge University Press, Cambridge, England.

Lakoff, George. 1976. Pronouns and Reference. In James McCawley, ed.,
{\it Syntax and Semantics}, Vol.7, Academic Press, New York, NY,
275--335.

Lewis, David. 1979. Scorekeeping in a Language Game. {\it Journal of
Philosophical Logic}, 8, 339--359.

Lujan, Marta. 1985. Stress and Binding of Pronouns. {\it Papers of the
21st Regional Meeting}, Chicago Linguistic Society, Chicago
University, 248--262.

Nakatani, Christine. 1993. Accenting on Pronouns and Proper Names in
Spontaneous Narrative. In {\it Proceedings of the ESCA Workshop on
Prosody}, Lund, Sweden. 

Prevost, Scott. 1996. An Information Structural Approach to Spoken
Language Generation. 
In {\it Proceedings of the 34th Meeting of the
Association of Computational Linguistics}, Santa Cruz, CA, 294--301.

Prince, Ellen. 1981. Toward a Taxonomy of Given--New Information. In
{\it Radical Pragmatics}, Academic Press, New York, NY, 223--255.

Pruest, Hub. 1992. {\it On Discourse Structuring, VP Anaphora and
Gapping.}  Ph.D. Thesis. University of Amsterdam.

Rooth, Mats. 1992. A Theory of Focus Interpretation. {\it Natural
Language Semantics} 1(1), 75--116.

Rooth, Mats. 1993. A Hybrid Architecture for Focus. An invited lecture
at the Nineth Amsterdam Colloquium.

Sgall, Petr, Eva Haji\v{c}ov\'{a}, and Jarmila
Panevov\'{a}. 1986. {\it The Meaning of the Sentence in its Semantic
and Pragmatics Aspects}, Reidel, Dordrecht and Academia, Prague.

Sidner, Candy. 1979. {\it Towards a Computational Theory of Definite
Anaphora Comprehension in English Discourse.} Ph.D. Thesis, Technical
Report 537, Artificial Intelligence Laboratory, MIT.

Sidner, Candace. 1983.  Focusing in the Comprehension of Definite
Anaphora.  In Brady, M. and R. Berwick, eds., {\em Computational
Models of Discourse}, The MIT Press, Cambridge, MA, 267--330.

Williams, Edwin. 1980. Remarks on Stress and Anaphora. {\it Journal of
Linguistic Research}, 1(3).

\endgroup
\end{footnotesize}

\vspace{1cm}
\noindent
Megumi Kameyama\\
{\em AI Center and CSLI, SRI International}\\
{\em 333 Ravenswood Avenue, Menlo Park, CA 94025, U.S.A.}\\
{\em e--mail: megumi\verb+@+ai.sri.com}

\end{document}